\documentclass{article}
\usepackage{spconf,amsmath,graphicx}
\usepackage{multirow}
\usepackage{booktabs,subcaption}
\usepackage{amsmath,amsfonts,dsfont,hyperref}

\title{LEARNING TO CHARACTERIZE ADVERSARIAL SUBSPACES}
\name{Xiaofeng Mao\sthanks{Contacted e-mail: mxf164419@alibaba-inc.com}, Yuefeng Chen, Yuhong Li, Yuan He, Hui Xue}
\address{Alibaba Group, China}
\begin{document}
\ninept
\maketitle
\begin{abstract}

Deep Neural Networks (DNNs) are known to be vulnerable to the maliciously generated adversarial examples. To detect these adversarial examples, previous methods use artificially designed metrics to characterize the properties of \textit{adversarial subspaces} where adversarial examples lie. However, we find these methods are not working in practical attack detection scenarios. Because the artificially defined features are lack of robustness and show limitation in discriminative power to detect strong attacks. To solve this problem, we propose a novel adversarial detection method which identifies adversaries by adaptively learning reasonable metrics to characterize adversarial subspaces. As auxiliary context information, \textit{k} nearest neighbors are used to represent the surrounded subspace of the detected sample. We propose an innovative model called Neighbor Context Encoder (NCE) to learn from \textit{k} neighbors context and infer if the detected sample is normal or adversarial. We conduct thorough experiment on CIFAR-10, CIFAR-100 and ImageNet dataset. The results demonstrate that our approach surpasses all existing methods under three settings: \textit{attack-aware black-box detection}, \textit{attack-unaware black-box detection} and \textit{white-box detection}.
\end{abstract}

% NCE borrows the idea of Transformer and uses multi-head self attention to imitate the relation among neighbors and detected samples.

%

\section{Introduction}
\label{sec:intro}
Deep Neural Networks (DNNs) have been widely employed in a large variety of applications such as image classification~\cite{krizhevsky2012imagenet,he2016deep}, speech recognition~\cite{graves2013speech}, and natural language processing~\cite{vaswani2017attention}. However, recent works~\cite{szegedy2013intriguing,goodfellow2014explaining} have found that DNNs are not robust to \textit{adversarial examples}: samples added with some imperceptible perturbations but misleading a well trained model to output wrong predictions. It arouses a great concern about the potential unsafety of DNN applications~\cite{song2018physical,komkov2019advhat}. Thus, understanding and characterizing the behavior of adversarial examples has become an imperative research topic, as it helps to design better model protection mechanisms. 
% In this paper, we study the adversarial examples' abnormal behavior of internal representation in model, and differentiate them from normal samples. 

The existence of adversarial examples can be attributed to some unavoidable regions in high dimensional representation space, within which all the points have low probability and mislead the classifier to output some results totally contrary to human understanding. These regions, also known as \textit{adversarial subspaces}~\cite{tramer2017space}, have been extensively studied in many previous works~\cite{ma2018characterizing,fawzi2016robustness,liu2019detection}. Aiming to characterize adversarial regions, the earliest work~\cite{feinman2017detecting} uses Kernel Density (KD) as a measure since adversarial subspaces usually have lower probability density.~\cite{ma2018characterizing} follows up this work, they find KD is not effective enough in more complicated situation, and propose a more discriminative metric using Local Intrinsic Dimensionality (LID)~\cite{houle2017local}. LID is based on classical expansion models theory~\cite{houle2012generalized} and the hypothesis that the expansion dimensions of local adversarial regions are often higher than normal regions. 
\begin{figure}[!htb]
\centering
\includegraphics[width=8.0cm]{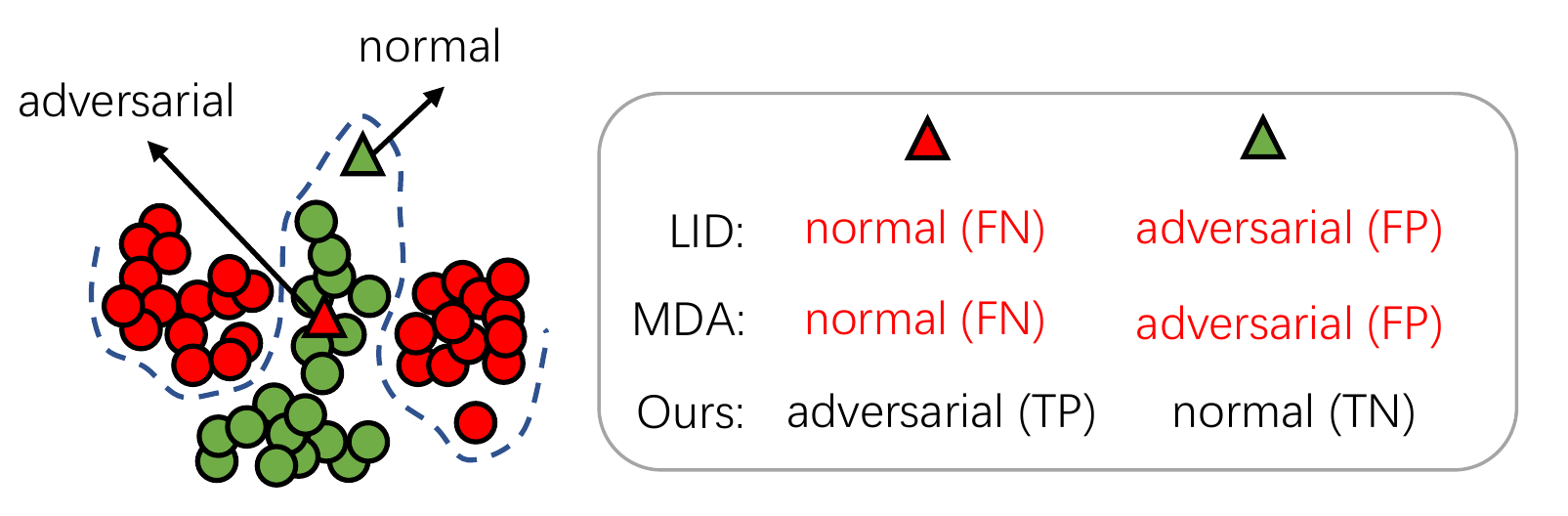}

\caption{A failure case of adversarial detection methods of LID and MDA, which demonstrates how false positive (FP) and false negative (FN) samples appear. While our method can avoid the occurrence of this successfully.}
\label{fig:intro}
\end{figure}
Instead of focusing local subspaces, another kind of methods use the distance to each class center of features to detect adversarial examples.~\cite{carrara2018adversarial} trains an LSTM network on a sequence of such distance to find adversarial examples. More recently,~\cite{lee2018simple} does Mahalanobis Distance Analysis (MDA) on adversarial example to each class conditional Gaussian distributions estimated by Gaussian Discriminant Analysis (GDA). MDA achieves the state of the art by improving baseline of the adversarial detection with a big margin.

However, all existing detection methods featurize the adversarial subspaces using artificially designed metrics merely. Despite elegant results have been achieved on some simple benchmarks, e.g., MNIST, we still find these metrics are weak and unstable under more powerful attacks or more complex data distribution. Specially, Fig.~\ref{fig:intro} shows how LID~\cite{ma2018characterizing} and MDA~\cite{lee2018simple} fail to detect adversarial examples in disordered embedding space. Samples in red and green represent two different categories and the dash line indicates decision boundary of the model. In the figure, the red triangle is a malicious sample but still on the data manifold, while the green triangle is normal but seems far away from the sample distribution. Both LID and MDA metrics confidently recall the green triangle as a adversarial example, with letting the real adversary off. This case reveals that even advanced LID and MDA metrics also have limitation in detection of more confusing adversaries.

In this paper, we propose a new idea that using a parameterized model to adaptively learn deep features for detecting adversaries. Different from existing methods which decoupled the adversarial detection into two stages: artificially characterizing the samples and classifying based on the obtained features, our method combines the two stages into one end-to-end learning framework. Therefore, our learned metric is more proper, and has stronger generalization and representation power for classifying adversarial examples compared with previous hand-designed metrics. Motivated by \cite{feinman2017detecting}, we uses the \textit{k} nearest neighbors to quantify the surrounded subspace of the detected sample. Generally, neighbors in the same subspace can form a topology. Every node (neighbor) in topology has their own attributes, and correlates to each other. If the surrounding neighbor points are sufficiently dense, the properties of these topologies can also reflect the characters of subspaces in some respects. Based on this recognition, we propose a novel Neighbor Context Encoder (NCE). For a specified detected sample, NCE learns the feature from the topology of its \textit{k} nearest neighbors and infers if the detected sample is normal or adversarial. We borrow Transformer~\cite{vaswani2017attention} to encode a sequence of input neighbor points, which is ranked by their distance to the detected sample in descending order. The attribute of each neighbor point consists of three parts: 1) the distance to detected sample; 2) the class label and 3) the position in input sequence, as shown in Fig.~\ref{fig:archi}. Then, the Transformer encoder effectively imitate the relation among neighbors and detected samples, using multi-head self attention. 
% Unlike the Transformer, NCE only takes the start position of the output as final prediction, which means the probability of the detected sample being a clean or adversarial example. We note that our method is different from~\cite{metzen2017detecting}, which augments neural networks with a small “detector” subnetwork and learn to distinguish genuine and adversarial data without thinking the space characteristics.
We conduct thorough experiment to analyze the proposed method under three settings: \textit{attack-aware black-box detection}, \textit{attack-unaware black-box detection} and \textit{white-box detection}. The result shows that our method achieves state of the art detecting FGSM~\cite{goodfellow2014explaining}, BIM~\cite{kurakin2016adversarial}, Deep-Fool~\cite{moosavi2016deepfool} and C\&W~\cite{carlini2017towards} attacks on CIFAR-10~\cite{krizhevsky2009learning}, CIFAR-100~\cite{krizhevsky2009learning} and ImageNet~\cite{deng2009imagenet} datasets. 

Our contributions can be summarized as follows:

(1) We propose the first end-to-end learning framework for characterizing adversarial subspaces, which greatly promote the traditional hand-designed descriptor in detection of adversarial examples.

(2) We propose a novel Neighbor Context Encoder (NCE) to learns the feature of the subspaces formed by surrounded \textit{k} nearest neighbors.

(3) Extensive experiments
illustrate that our method gains superior performance on multi-attacks and generalizes well on other more powerful attacks.

% For detail introduction of each part please refer to section \ref{ssec:Learning}. Transformer encoder is suitable for this problem because it effectively learns the relation among neighbors and detected samples, using multi-head self attention. Unlike the Transformer, NCE only takes the start position of the output as final prediction, which means the probability of the detected sample being a clean or adversarial example. We note that our method is different from~\cite{metzen2017detecting}, which augments neural networks with a small “detector” subnetwork and learn to distinguish genuine and adversarial data without thinking the space characteristics.

% The inspiration comes from the domain of Topological Data Analysis (TDA). Generally, samples in the same subspace can form a complex network topology. They have their own attributes, and correlate to each other. Using Fig \ref{fig:intro} for visualization, each sample can unite their surrounded samples and constitute a topology. If the surrounding sample points are sufficiently dense, the properties of these topologies can also reflect the characters of subspaces in some respects. Therefore, it makes sense that we propose learning  

\section{PROPOSED METHOD}
\label{sec:method}
\begin{figure}[!htb]
\includegraphics[width=8.6cm]{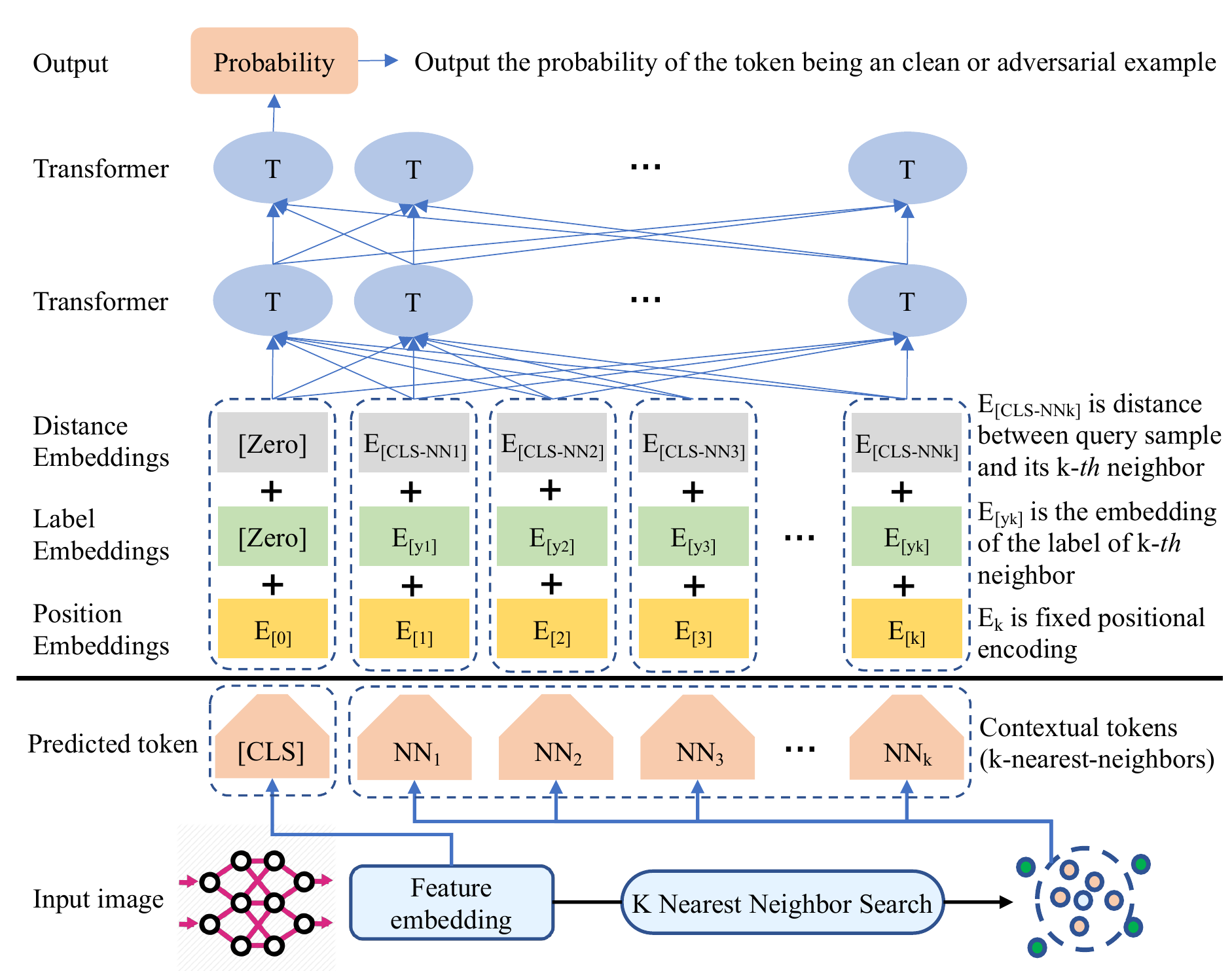}
\caption{The overall network architecture of NCE.}
\label{fig:archi}
\end{figure}

We introduce our method and its detailed implementation in this section. Our insight is parametrizing a DNN for learning to characterize local adversarial regions. However, traditional neural networks fail to process these region inputs because they lie in high dimensional feature space and we cannot quantify them. Fortunately, motivated by~\cite{feinman2017detecting}, which gives a hypothesis that adversarial regions can be presented by locally surrounded nearest neighbors, we regard adversarial subspace as a sequence of \textit{k} nearest neighbors. The goal is to predict the normality or abnormality of the detected sample based on its surrounded adversarial subspace.
We reformulate this problem as a node/token prediction task, where detected sample is regarded as predicted token, and its neighbors are viewed as contextual tokens.  Detailly, to predict an example being normal or adversarial, we first search its \textit{k} nearest neighbors in feature space. Every neighbor is treated as a token and connects each other to form a sequence with the length of \textit{k}. Detected sample is added in the start position as a special classification token (\verb [CLS] ). Then we use proposed NCE to encode the input sequence and output the probability of predicted sample (i.e.\verb [CLS] token) being normal or adversarial. We note that our method is different from~\cite{metzen2017detecting}, which augments neural networks with a small “detector” subnetwork and learn to distinguish genuine and adversarial data without considering the space characteristics. General workflow and our model architecture are shown in Fig.~\ref{fig:archi}. 

\subsection{Input preprocess}
\label{ssec:Input preprocess}
More details about how to construct the training data will be introduced in this section. Suppose that we have a pre-trained target model $H(x)$, which was not defended yet. Training images $X_{cln}$ are used for generating adversarial samples $X_{adv}$ and noisy samples $X_{nis}$. Here noisy examples are generated by adding random noise to clean samples $X_{cln}$. Adding noisy images in training set as negative samples is a data augmentation trick first proposed in~\cite{feinman2017detecting,carlini2017adversarial}. By feeding samples into model $H(x)$, we get three sets of $l$-th layer feature output $\left\{f_{X_{cln}}^{l},f_{X_{nis}}^{l},f_{X_{adv}}^{l}\right\}$ and corresponding label $\left\{Y_{cln}=0,Y_{nis}=0,Y_{adv}=1\right\}$, where 0,1 present negtive and positive samples respectively.

In addition to the training images, we also use a query-purpose image library $X^{c}=\left\{x_{1}^{c}, x_{2}^{c},..., x_{n}^{c}\right\}$ and its corresponding label $Y^{c}\in \left\{ 1,2,..,C\right\}$, $C$ is the number of classes. So that the training images can query its \textit{k} nearest neighbors from this database. $X^{c}$ are also fed into the model $H(x)$ for obtaining the feature set $f_{X^{c}}^{l}=\left\{f_{x_{1}^{c}}^{l}, f_{x_{2}^{c}}^{l},..., f_{x_{n}^{c}}^{l}\right\}$. For each $f^{l}_{x}\in\left\{f_{X_{cln}}^{l},f_{X_{nis}}^{l},f_{X_{adv}}^{l}\right\}$, we calculate the distance to the feature set $f_{X^{c}}^{l}$ and retrieve the nearest \textit{k} images as :

\begin{equation}
\begin{split}
    \mathrm{NN}_{1}, \mathrm{NN}_{2}, ..., \mathrm{NN}_{k} &= topk([d(f^{l}_{x}, f^{l}_{x^{c}_{1}}), d(f^{l}_{x}, f^{l}_{x^{c}_{2}}),\\
    &...,d(f^{l}_{x}, f^{l}_{x^{c}_{n}})])
\end{split}
\end{equation}
where $topk(\cdot)$ finds \textit{k} samples in $X^{c}$ with smallest distance, and re-ranks them with descending order, then returns their indexes $\mathrm{NN}_{1}, \mathrm{NN}_{2}, ..., \mathrm{NN}_{k}$. $d(\cdot )$ is a distance function, we use Euclidean distance here. Query-purpose image library $X^{c}$ provides contextual information for classifying detected sample. It is remarkable that $X^{c}$ can be same as training images $X_{cln}$. We found the model also achieved good performance when using $X_{cln}$ as $X^{c}$ directly in experiment. 

% Regarding $\mathrm{NN}_{1}, \mathrm{NN}_{2}, ..., \mathrm{NN}_{k}$ as contextual word tokens in a sentence, and detected sample as a classified word token(\verb [CLS] ), this problem is equivalent with the Masked LM task~\cite{devlin2018bert} in Natural Language Processing (NLP).

\subsection{Learning to characterize adversarial subspaces}
\label{ssec:Learning}
We borrow the idea of Transformer~\cite{vaswani2017attention} to encode token representations. For a given token, its input representation is constructed by summing the corresponding distance, label, and position embeddings. A visualization of this construction can be seen in Fig.~\ref{fig:archi}. Label embedding presents the label information of neighbors. We assume the label of neighbors as $y_{\mathrm{NN}_{i}}^{c} \in Y^{c}$, which is one-hot encoded. The learnable label embedding is created by embeddings lookup:
\begin{equation}
    \mathrm{E}_{[y_{i}]} = W^{T} \times y_{\mathrm{NN}_{i}}^{c}
\end{equation}
where $W \in \mathbb{R}^{C\times D}$ is embedding parameters, $D$ is the embedding size. Distance embedding presents the computed distance of neighbor to the detected sample:
\begin{equation}
    \mathrm{E}_{[\mathrm{CLS}-\mathrm{NN}_{i}]} = d(f^{l}_{x}, f^{l}_{x^{c}_{\mathrm{NN}_{i}}})
\end{equation}
Since the distance is a scalar, we replicate it to the same dimension size with the label embeddings. Position embedding injects some information about the relative or absolute position of tokens in the sequence position. We use sine and cosine functions of different frequencies as fixed position embedding as in~\cite{vaswani2017attention}:
\begin{equation}
    \mathrm{E}_{[i]} = [\sin(\frac{i}{t^{0}}),\cos(\frac{i}{t^{\frac{1}{D}}}),...,((D-1) \bmod 2)\cos(\frac{i}{t^{\frac{D-1}{D}}})]
\end{equation}
where $t$ is the wavelengths parameter, here we set $t=10000$. Finally, our input embedding sequence is the element-wise sum of $\mathrm{E}_{[\mathrm{CLS}-\mathrm{NN}_{i}]}$, $\mathrm{E}_{[y_{i}]}$ and $\mathrm{E}_{i}$:
\begin{equation}
 \mathrm{E}_{[\mathrm{NN}_{i}]}= \underbrace{\mathrm{E}_{[\mathrm{CLS}-\mathrm{NN}_{i}]}}_{\text{distance emb}}+\underbrace{\mathrm{E}_{[y_{i}]}}_{\text{label emb}}+\underbrace{\mathrm{E}_{[i]}}_{\text{position emb}}
 \label{eq:5}
\end{equation}
For $\mathrm{E}_{[\mathrm{CLS}]}$ there is no label or distance embedding available, so we use zero vectors instead. Subsequently, some Transformer layers process the input and predict the probability of the\verb [CLS] token being an normal or adversarial example:
\begin{equation}
\begin{aligned}
 & p_{[\mathrm{CLS}]},p_{[\mathrm{NN}_{1}]},p_{[\mathrm{NN}_{2}]},...,p_{[\mathrm{NN}_{k}]} =
 \\ Transf&ormers(\mathrm{E}_{[\mathrm{CLS}]}, \mathrm{E}_{[\mathrm{NN}_{1}]},\mathrm{E}_{[\mathrm{NN}_{2}]},...,\mathrm{E}_{[\mathrm{NN}_{k}]})
\end{aligned}
\end{equation}
where $p_{[\mathrm{CLS}]},p_{[\mathrm{NN}_{1}]},p_{[\mathrm{NN}_{2}]},...,p_{[\mathrm{NN}_{k}]}$ is the output probabilities of the Transformer corresponding to each input token. We only use the prediction $p_{[\mathrm{CLS}]}$ of\verb [CLS] token in this case. During training, we use $\left\{X_{cln}, X_{nis}, X_{adv}\right\}$ and corresponding labels $\left\{Y_{cln}=0,Y_{nis}=0,Y_{adv}=1\right\}$ to train the parameters of transformer layers and the label embeddings. 

\section{EXPERIMENT}
\label{sec:exp}
In this section, we present the experimental results for detecting the adversarial examples crafted on CIFAR-10~\cite{krizhevsky2009learning}, CIFAR-100 and large-scale ImageNet~\cite{deng2009imagenet} visual recognition datasets. We compare the detection capability of our method with two strong baseline: LID~\cite{ma2018characterizing} and MDA~\cite{lee2018simple}. In the experiment, four attack methods: FGSM~\cite{goodfellow2014explaining}, BIM~\cite{kurakin2016adversarial}, DeepFool~\cite{moosavi2016deepfool} and C\&W~\cite{carlini2017towards} are adopted as adversaries.

Since the way to attack a model can be various under different settings. We simulate three types of detection scenarios to test the proposed method. i) the first and simplest scenario is \textit{attack-aware black-box detection}, in which we knows the attack scheme and can use generated adversarial examples to train the detector. ii) another situation is \textit{attack-unaware black-box detection}. It is more practical since we usually have no information about what strategy the attacker used. iii) the last and most difficult is \textit{white-box detection}. We assume that attacker knows the scheme and parameters of the detector, and can design special
methods to cheat the detection model. 
\subsection{Attack-aware black-box detection}
\label{ssec:attack-aware bbox det}

In this case, we train ResNet~\cite{he2016deep}, DenseNet~\cite{huang2017densely} and ShuffleNet~\cite{ma2018shufflenet} for classifying CIFAR-10 and CIFAR-100 images. And for larger ImageNet benchmark, pretrained AlexNet and ResNet are employed. 10000 test images in CIFAR-10\&100 and 10000 randomly selected images in validation set of ImageNet (10 images per category) are used as test datasets. To prevent the overlap, our detector is trained and evaluated on test datasets. As suggested in~\cite{feinman2017detecting,carlini2017adversarial}, we augment noisy samples which add random Gaussian noise on clean images as negative training data. For the Transformer, we set the dimension size of both label and position embedding as 16. Since the distance is a continuous number, we replicate it to the same dimension size with the label and position embedding. In the experiment, we use multi-layer output as features and search the \textit{k} nearest neighbors based on the feature of each layer. Then we concatenate the embeddings computed by Eq.~\ref{eq:5} on each layer as the input. The designed detection network is relatively small because we find that greater capacity does not help for the efficiency. We only use 1 Transformer encode layer, which contains a multi-head self attention layer with head number of 8, and subsequently 2 \textit{fc} layers with Dropout~\cite{srivastava2014dropout} of 0.1 probability.  

\begin{table*}[!htb]
\small
  \centering
  \caption{Comparison of attack-aware black-box detection performance using AUROC(\%) score. The best results are indicated in bold.}
  \begin{tabular}{cc|ccc|ccc|ccc|ccc}
    \toprule[1.2pt]
    \multirow{2}{*}{Dataset} &
    \multirow{2}{*}{Model} &
    \multicolumn{3}{c|}{FGSM(AUROC(\%))} & \multicolumn{3}{c|}{BIM(AUROC(\%))} &
    \multicolumn{3}{c|}{DeepFool(AUROC(\%))} & 
    \multicolumn{3}{c}{CW(AUROC(\%))}
    \\
    &  & LID & MDA & Ours & LID & MDA & Ours & LID & MDA & Ours & LID & MDA & Ours\\
    \midrule[0.7pt]
    \multirow{3}{*}{CIFAR10} & ShuffleNet & 93.65 & 99.36 & \textbf{99.73} & 98.02 & 98.68 & \textbf{98.74} & 79.24 & 67.26 & \textbf{84.94} & 72.64 & 62.16 & \textbf{82.89} \\
    & ResNet & 99.41 & 99.93 & \textbf{99.95} & 95.79 & 99.55 & \textbf{99.62} & 91.01 & 94.73 & \textbf{94.97} & 88.76 & 97.36 & \textbf{97.57} \\
    & DenseNet & 99.86 & 99.94 & \textbf{99.97} & 99.64 & \textbf{99.90} & 99.87 & 99.64 & 99.82 & \textbf{99.89} & 85.55 & 88.07 & \textbf{94.41} \\
    \midrule[0.7pt]
    \multirow{3}{*}{CIFAR100} & ShuffleNet & 92.57 & 99.30 & \textbf{99.83} & 92.37 & 96.92 & \textbf{97.62} & 66.37 & 73.35 & \textbf{84.35} & 61.71 & 69.77 & \textbf{80.39} \\
    & ResNet & 98.77 & 99.77 & \textbf{99.93} & 96.11 & 97.07 & \textbf{98.68} & 68.81 & 88.05 & \textbf{88.69} & 75.62 & 94.11 & \textbf{96.65} \\
    & DenseNet & 97.99 & 99.71 & \textbf{99.91} & 99.25 & 99.78 & \textbf{99.82} & 70.34 & 74.25 & \textbf{82.88} & 71.29 & 80.99 & \textbf{90.84} \\
    \midrule[0.7pt]
    \multirow{2}{*}{ImageNet} & ResNet & 92.96 & 98.21 & \textbf{99.55} & 93.33 & 97.82 & \textbf{99.56} & 96.45 & 96.56 & \textbf{97.86} & \textbf{72.30}  & 64.97  & 63.79 \\
    & AlexNet & 96.07 & 98.63 & \textbf{98.82} & 93.77 & 96.60 & \textbf{98.54} & 97.60 & 97.65 & \textbf{98.03} & 63.74 & 61.52 & \textbf{63.90} \\
    \bottomrule[1.2pt]
  \end{tabular}
  \label{tab:attack_aware}
\end{table*}

\begin{table*}[!htb]
\small
  \centering
  \caption{Comparison of AUROC(\%) under attack-unaware black-box detection setting. The best results are indicated in bold.}
  \begin{tabular}{cc|ccc|ccc|ccc|ccc}
    \toprule[1.2pt]
    \multirow{3}{*}{} &
    \multirow{3}{*}{} &
    \multicolumn{6}{c|}{ResNet} & \multicolumn{6}{c}{ShuffleNet}
    \\
    &  & \multicolumn{3}{c|}{DeepFool} & \multicolumn{3}{c|}{CW} & \multicolumn{3}{c|}{DeepFool} & \multicolumn{3}{c}{CW}
    \\
    &  & LID & MDA & Ours & LID & MDA & Ours & LID & MDA & Ours & LID & MDA & Ours
    \\
    \midrule[0.7pt]
    \multirow{2}{*}{ResNet} & FGSM & 72.45 & 78.90 & \textbf{88.70} & 84.59 & \textbf{95.22} & 84.50 & 71.47 & 73.27 & \textbf{84.87} & 65.23 & 72.65 & \textbf{86.72} \\
    & BIM & 83.98 & 88.72 & \textbf{88.81} & 78.86 & 93.72 & \textbf{94.60} & 57.76 & 63.26 & \textbf{80.58} & 52.77 & 67.15 & \textbf{84.31} \\
    \midrule[0.7pt]
    \multirow{2}{*}{ShuffleNet} & FGSM & 67.36 & 68.10 & \textbf{87.99} & 65.50 & 64.34 & \textbf{74.76} & 63.70 & 64.39 & \textbf{72.61} & 58.52 & 50.64 & \textbf{66.16} \\
    & BIM & 50.01 & 54.75 & \textbf{65.13} & 50.17 & 50.05 & \textbf{58.84} & 53.89 & 55.84 & \textbf{63.14} & 47.20 & 55.71 & \textbf{61.87} \\
    \bottomrule[1.2pt]
  \end{tabular}
  \label{tab:attack_unaware}
\end{table*}

Results are presented in Table \ref{tab:attack_aware}. we report the area under the receiver operating characteristic curve (AUROC) score as the metric for performance. Both two baseline methods have achieved high detection accuracy and AUROC score on gradient-based attacks (FGSM and BIM). Among them LID has worse performance because the feature of LID may not be discriminative enough in more complex and difficult adversarial detection tasks. Instead of calculating LID, MDA uses more robust mahalanobis distance as feature and get better AUROC score. Although those artificially defined features achieve great performance in detection of gradient-based attacks, for optimization-based attaks, they are weak and unstable. While our learned representations are effective in recall hard samples in detection tasks. It has higher AUROC on both four tested attacks. For CW attacker, we greatly promote the detection performance on CIFAR-10\&100 datasets, but on ImageNet, both MDA and our approach have lower AUROC because of the confusion among classes caused by a large number of categories.

\subsection{Attack-unaware black-box detection}
\label{ssec:attack-unaware bbox det}
As suggested in~\cite{ma2018characterizing}, for evaluating whether the model tuned on a simple attack can be generalized to detect other attacks, we give a generalization analysis. Here, we go a step further to test the detection ability on \textit{transfer attacks}~\cite{papernot2016practical}, e.g., performing attack on surrogate model and replaying it to the target model. ResNet and ShuffleNet trained on CIFAR10 dataset are used for detection generalizability test. As shown in Table \ref{tab:attack_unaware}, head of the row means training attacks and head of the column means testing attacks. For example, in third row first column of the table, 67.36\% presents the AUROC score when the LID detector is trained using seen FGSM attack on ShuffleNet and tested using unseen DeepFool attack on ResNet. The results demonstrate that both LID and MDA are not good at detecting unseen attacks in most cases. For instance, LID and MDA detector trained using BIM attack on ShuffleNet only get around 50\% AUROC score, which is equivalent to a random predictor. It implies that LID and MDA can only capture the characteristics of the seen attack, but not the characteristics of the common adversarial subspace. The main reason is that LID and MDA score may vary greatly on different attacks, which cause significant decline in classification. Instead, our approach gains huge improvement compared with the existing methods. Specifically, it can  enhance the baseline of detecting C\&W attack on ShuffleNet from 67.15\% to 84.31\% when model is trained using BIM attack on ResNet. Our model learns similar representation for different types of attacks, but varies greatly only between normal and adversarial examples. Thus, adversarial subspace can be characterized more exactly using our approach. 

\subsection{White-box detection}
\label{ssec:white-box}

Many adversarial detection methods are proved to be ineffective under adaptive attack.~\cite{carlini2017adversarial} incorporates kernel density into the adversarial objective and makes detection more difficult for the KD-based method. Similarly, we also combine LID and MDA as one of the optimizing objectives and fool the detector successfully. The fundamental cause of such detectors being vulnerable to adaptive attack is that they are differentiable. On the contrary, our model receives discrete sample point as input, which makes adaptive attack more difficult. Since it is hard to find an gradient-based adaptative attacker based on our method, we use the direction to a selected target sample to approximate the gradient, as in~\cite{sitawarin2019robustness}. We choose a nearest clean sample in $X_{cln}$ with different label as a target sample. The goal is moving the generated adversarial example closer to target sample in embedding space. In white-box case, we found only 16\% of these adversaries bypass our model, which shows good robustness on white-box attacks. We also conduct the same attack on LID and MDA, the fooling rate is up to 72\% and 80\% respectively.

\subsection{Ablation study}
\label{ssec:ablation}

\begin{figure}[!htb]
\begin{minipage}[b]{.5\linewidth}
  \centering
  \centerline{\includegraphics[width=4.0cm]{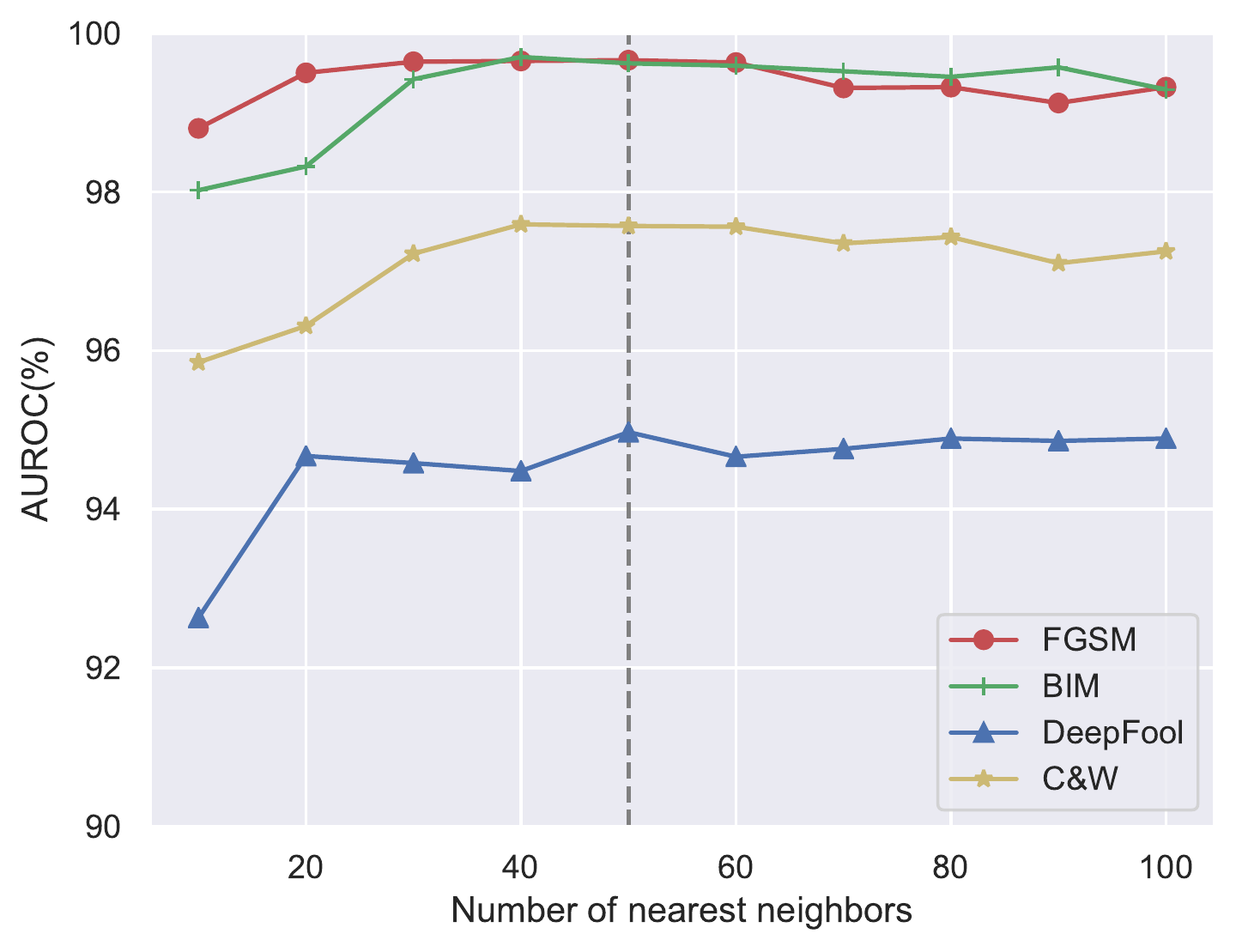}}
%  \vspace{1.5cm}
\end{minipage}
\hfill
\begin{minipage}[b]{0.5\linewidth}
  \centering
  \centerline{\includegraphics[width=4.0cm]{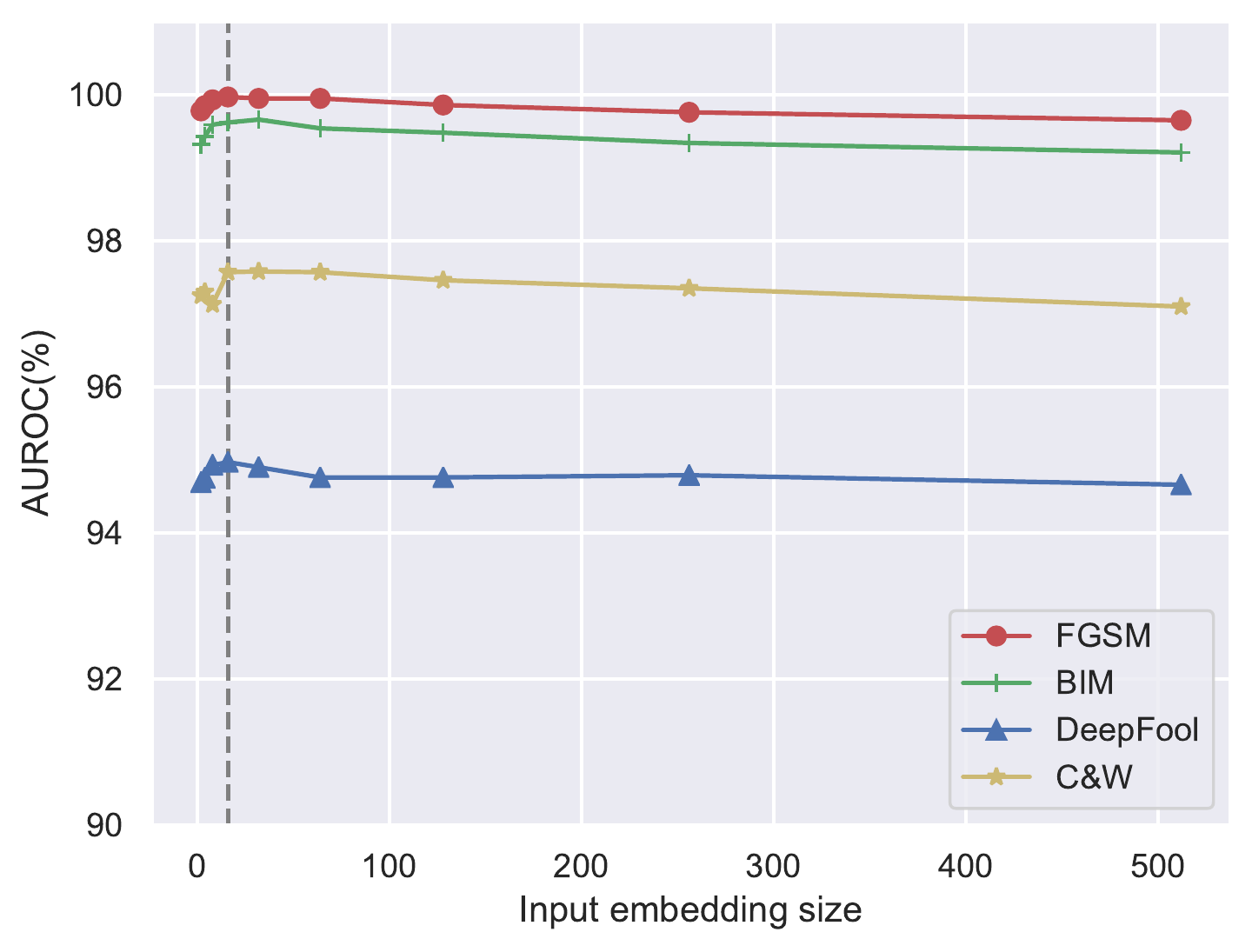}}
%  \vspace{1.5cm}
\end{minipage}
\caption{Performance plot of different choices of nearest neighbor number \textit{k} and input embedding size $D$.}
\label{fig:ablation}

\end{figure}

In the experiment, we found hyperparameters $k$ and $D$ playing an important role. We analyze how they affect the performance of model in Fig.~\ref{fig:ablation}. We draw the performance curve of AUROC(\%) corresponding different choices of nearest neighbor number \textit{k} and input embedding size $D$. In left-hand part, the AUROC score increases gradually with inputing more nearest neighbors. And we found the performance remaining unchanged when \textit{k} increases to 50. The performance draw of different input embedding size $D$ also have the same trend. It reaches the maximum peak when input embedding size equals to 16. This phenomenon shows that large embedding size or redundant neighbors are not helpful but confusing the detection of adversarial examples. So we set $k=50$ and $D=16$ as the empirical value in the experiments.

\vspace{-0.5cm}
\section{CONCLUSION}
\label{sec:conclusion}
In this paper, we study the problem of characterizing the properties of adversarial regions to detect adversarial examples. We use a parametrizd DNN named Neighbor Context Encoder (NCE) to adaptively learn reasonable metrics in characterization of adversarial subspaces. The experimental results suggest our method is effective in detection of four classic adversaries: FGSM, BIM, Deep-Fool and C\&W. It exceeds all the existing adversarial detection methods in both accuracy and generalization ability. Besides, our method is the first simple solution of using parametrizd model to learn features of the adversarial subspace. We think there will be many potential future works in this direction.

In the training stage, our model only accepts the sequence of \textit{k} nearest neighbors, where each neighbor are presented as the feature of each layer. It does not care the feature evolution throughout the forward pass in the model. Learning the correlation of the internal features is a future work.

Another future direction is designing a strong attacker against our nondifferentiable model. An adaptative attacker also helps to build robust detection models.

\vfill\pagebreak

% \section{REFERENCES}
% \label{sec:refs}

% List and number all bibliographical references at the end of the
% paper. The references can be numbered in alphabetic order or in
% order of appearance in the document. When referring to them in
% the text, type the corresponding reference number in square
% brackets as shown at the end of this sentence \cite{C2}. An
% additional final page (the fifth page, in most cases) is
% allowed, but must contain only references to the prior
% literature.

% References should be produced using the bibtex program from suitable
% BiBTeX files (here: strings, refs, manuals). The IEEEbib.bst bibliography
% style file from IEEE produces unsorted bibliography list.
% -------------------------------------------------------------------------
\bibliographystyle{IEEEbib}
\bibliography{strings,refs}

\end{document}